\documentclass[conference]{IEEEtran}
\ifCLASSINFOpdf
\else
\fi

\usepackage{cite}
\usepackage{amsmath,amssymb,amsfonts}
\usepackage{graphicx}
\usepackage{textcomp}
\usepackage{xcolor}
\usepackage{float}
\usepackage[ruled,vlined]{algorithm2e}
\usepackage{algpseudocode}
\usepackage{makecell}
\usepackage{caption}
\usepackage{subfigure}
\usepackage{mdwlist}
\usepackage{enumitem} 
\usepackage{mathrsfs}
\usepackage{booktabs}

\begin{document}
%
\title{Hardware Acceleration of Explainable Machine Learning using Tensor Processing Units}

\author{\IEEEauthorblockN{Zhixin Pan and Prabhat Mishra}
\IEEEauthorblockA{Department of Computer \& Information Science \& Engineering\\
University of Florida, Gainesville, Florida, USA}}


%


\maketitle

\begin{abstract}
Machine learning (ML) is successful in achieving human-level performance in various fields. However, it lacks the ability to explain an outcome due to its black-box nature. While existing explainable ML is promising, almost all of these methods focus on formatting interpretability as an optimization problem. Such a mapping leads to numerous iterations of time-consuming complex computations, which limits their applicability in real-time applications. In this paper, we propose a novel framework for accelerating explainable ML using Tensor Processing Units (TPUs). The proposed framework exploits the synergy between matrix convolution and Fourier transform, and takes full advantage of TPU's natural ability in accelerating matrix computations. Specifically, this paper makes three important contributions. (1) To the best of our knowledge, our proposed work is the first attempt in enabling hardware acceleration of explainable ML using TPUs. (2) Our proposed approach is applicable across a wide variety of ML algorithms, and effective utilization of TPU-based acceleration can lead to real-time outcome interpretation. (3) Extensive experimental results demonstrate that our proposed approach can provide an order-of-magnitude speedup in both classification time (25x on average) and interpretation time (13x on average) compared to state-of-the-art techniques. 
\end{abstract}


%
\IEEEpeerreviewmaketitle

\section{Introduction}




Machine learning (ML) techniques powered by deep neural networks (DNNs) are pervasive across various application domains. Recent advances in ML algorithms have enabled promising performance with outstanding flexibility and generalization. Unfortunately, ML is not able to interpret the outcome (e.g., explain its prediction) since it produces the outcome based on computations inside a ``black-box''. This lack of transparency severely limits the applicability of ML. For example, in case of safety-critical applications, a user would like to know the rationale behind its decision to trust its prediction. Similarly, an ML algorithm is expected to provide an outcome (prediction) as well as interpretation of the outcome in many security applications so that a security expert can act judiciously. For example, during malware detection, it is important to know whether a software is malicious or benign. In order to trust the results, it is important to understand the rationale for such a classification. Moreover, the interpretation of the results is crucial to enable the localization (e.g., clock cycle and specific reason) of the malicious activity in a malware \cite{iccd20}.  
 

Explainable ML is a promising direction to enable outcome interpretation. There are a large number of existing efforts in the area of explainable ML ~\cite{EML}. Due to inherent inefficiency in these algorithms, they are not applicable in real-time systems. These algorithms treat the explanation process as an extra procedure, and performs the interpretation outside the learning model, which makes them inefficient in practice. Specifically, it solves a complex optimization problem that consists of numerous iterations of time-consuming computations. As a result, such time-consuming interpretation is not suitable for time-sensitive applications with soft or hard deadlines. In soft real-time systems, such as multimedia and gaming devices, inefficient interpretation can lead to unacceptable Quality-of-Service (QoS). In hard real-time systems, such as safety-critical systems, missing task deadlines can lead to catastrophic consequences.

In this paper, we propose an efficient framework to perform explainable ML utilizing  Tensor Processing Unit (TPU). TPU is an Application Specific Integrated Circuit (ASIC) developed specifically to accelerate the computations in deep neural networks. TPU has shown great potential in improving training efficiency of various ML tasks~\cite{TPU1,TPU2,TPU3,TPU4,TPU5}. According to the experimental evaluation in~\cite{intro}, TPU can achieve 15 to 30 times higher throughput compared to contemporary CPU and GPU based acceleration. \textit{\textbf{Our proposed approach effectively utilizes the synergy between matrix based representation of interpretation procedure and TPU-based acceleration of matrix operations}}. Specifically, this paper makes the following three important contributions.

\begin{enumerate}
    \item To the best of our knowledge, our approach is the first attempt in TPU-based hardware acceleration of  explainable machine learning. 
    \item Our proposed method exploits inherent advantage of TPU in transforming a complex interpretation process to a simple computation equivalent to one forward pass.
    \item Experiments using two popular ML models demonstrate that our proposed approach can provide significant improvement compared to the state-of-the-art methods. 
\end{enumerate}

The rest of this paper is organized as follows. We survey related efforts in Section~\ref{relwork}. Section~\ref{proposed} describes our proposed method for accelerating explainable ML algorithms using TPU. Section~\ref{exp} presents experimental results. Finally, Section~\ref{conclude} concludes the paper.

\section{Background and Related Work}
\label{relwork}

\subsection{Accelerating Machine Learning using TPU}
\begin{figure}[htbp]
\centering
\vspace{-0.1in}
\includegraphics[scale=0.29]{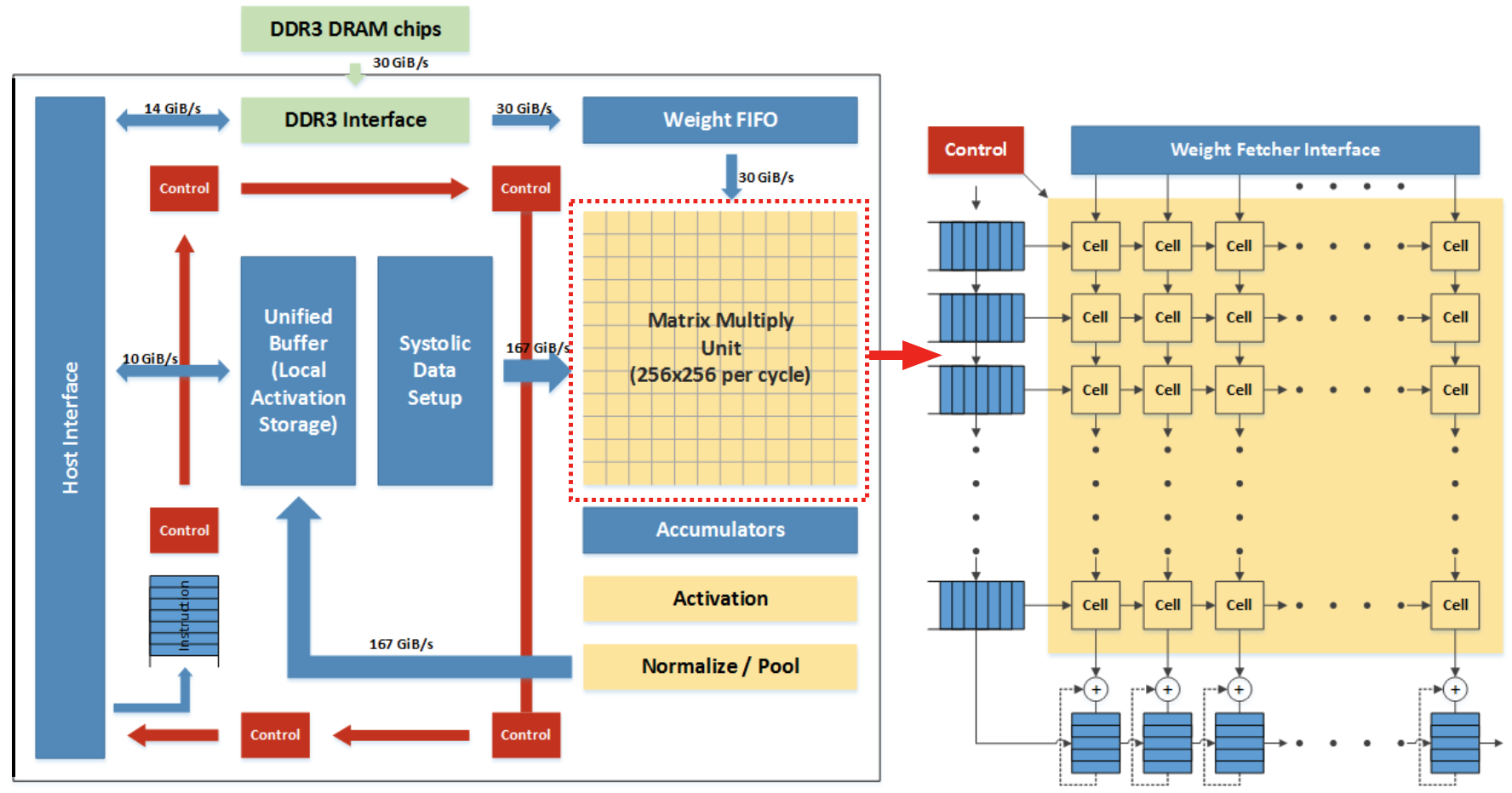}
\caption{The structure of TPU, whose key component is the Matrix Multiply Unit (MXU) implemented by systolic array.}
\label{fig:structure}
\vspace{-0.1in}
\end{figure}

Tensor Processing Unit (TPU) proposed by Google is a domain-specific hardware for accelerating computation process of deep learning models. There are two fundamental reasons for the superior performance of TPUs compared to CPU/GPU based acceleration:  \textit{quantization} and \textit{systolic array}~\cite{SysArry}. Quantization is the first step of optimization, which uses 8-bit integers to approximate 16-bit or 32-bit floating-point numbers. This can reduce the required memory capacity and computing resources. Systolic array is a major contributor to TPU's efficiency due to its natural compatibility with matrix manipulation coupled with the fact that computation in neural networks can be represented as matrix operations. Figure~\ref{fig:structure} shows an overview of the simplified structure of TPU.

From a global perspective, the core of the entire TPU is the Matrix Multiply Unit, which is a 256$\times$256 systolic array composed of multiple computation cells. Each cell receives a weight parameter along with an input signal at a time, and performs accumulation of their products. Once all weights and input signals are propagated to neighboring cells, top to bottom and left to right respectively, it immediately starts next round of iteration. By this calculation scheme, the entire matrix multiplication is actually completed by the collaboration of all computation cells. The systolic array of MXU contains 256 × 256 = 65,536 ALUs, which means that the TPU can process 65,536 8-bit integer multiplications and additions per cycle. Due to the systolic architecture, input data is actually reused for multiple times. Therefore, it can achieve higher throughput while consuming less memory bandwidth. \textit{While TPU has been successfully used for accelerating machine learning algorithms, there are no prior efforts in utilizing TPU for accelerating explainable ML}.

\subsection{Explainable Machine Learning}
The demand for explainable ML has been steadily increasing ever since ML algorithms were widely adopted in many fields, especially in security domains. In this section, we briefly introduce explainable ML, and then discuss one widely adopted explanation technique. In general, explainable ML seeks to provide interpretable explanation for the results of machine learning model. Specifically, given an input instance $\bf x$ and a model $M$, the classifier will generate a corresponding output $\bf y$ for $\bf x$ during the testing time. Explanation techniques then aim to illustrate why instance $\bf x$ is transformed into $\bf y$. This often involves identifying a set of important features that make key contributions to the forward pass of model $M$. If the selected features are interpretable by human analysts, then these features can offer an “explanation”. Most existing explanation methods exploit the concept of ``model distillation''~\cite{ModelDistillation}. The basic idea of model distillation is that it develops a separate model called as ``distilled model'' to be an approximation of the input-output behavior of the target machine learning model. This distilled model, denoted as $M^*$, is usually chosen to be inherently explainable by which a user is able to identify the decision rules or input features influencing the outputs. In reality, model distillation is composed of three major steps.
\begin{itemize}
    \item \textbf{\textit{Model Specification:}} To begin with, the type of distilled model has to be specified. This often involves a challenging trade-off between transparency and expression ability. A complex model can offer better performance in mimicking the behavior of original model $M$. But meanwhile, increasing complexity also lead to the inevitable drop of model transparency, where the distilled model itself becomes hard to explain, and vice versa.
    \item \textbf{\textit{Model Computation:}} Once the type of distilled model $M^*$ is determined and corresponding input-output pairs are provided, the model computation task aims at searching for optimal parameters $\theta$ for $M$ using Equation~\ref{eqn:theta}, which is an optimization problem.  

    \vspace{-0.15in}
    \begin{equation}
    \theta = \arg\min\limits_{\bf {\theta}}||M^*({\bf x}) -\textbf{y}|| 
    \label{eqn:theta}
    \vspace{-0.1in} 
    \end{equation}
    
    \item \textbf{\textit{Outcome Interpretation:}} Based on the computed distilled model in the previous step, the explanation boils down to measuring the contribution of each input feature in producing the classification result.  For instance, assume a linear regression model is applied which can always be expressed as a polynomial. Then by sorting the terms with amplitude of coefficients, we have access to many crucial information including the input features it found to be most discriminatory, and features or output correlations relevant for classification. 
\end{itemize}


Notice that interpreting the distilled model may not provide us with deep insights into the internal representation of the ML model, or demonstrate anything about the model's learning process, but it can at least provide insight into the correlations and relational rules to  explain how the ML model makes a decision. \textit{While explainable machine learning has received significant attention in recent years, to the best of our knowledge, there are no prior efforts in enabling hardware acceleration of explainable machine learning}. The next section describes our proposed approach for TPU-based acceleration of explainable machine learning.

\section{TPU-based Explainable Machine Learning}\label{proposed}

\subsection{Overview}
\begin{figure}[htbp]
\centering
\vspace{-0.15in}
\includegraphics[scale=0.36]{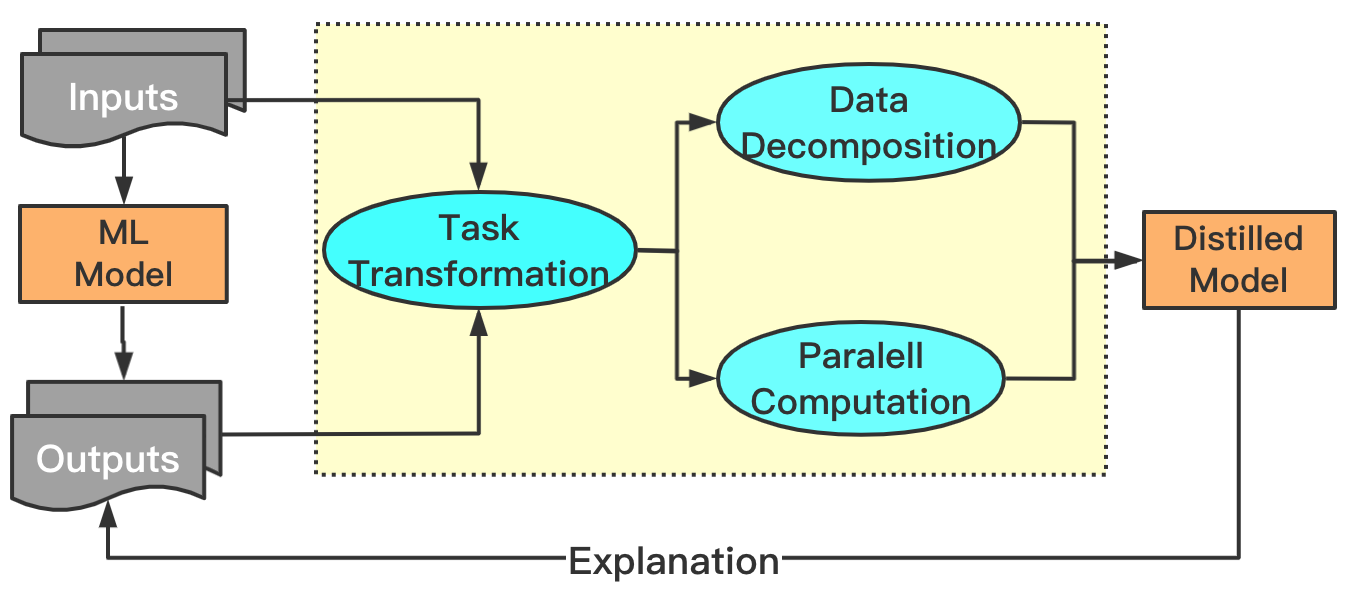}
\vspace{-0.1in}
\caption{Our proposed TPU-based hardware acceleration framework consists of three major activities: task transformation, data decomposition and parallel computation.}
\label{fig:overview}
\vspace{-0.1in}
\end{figure}

Figure~\ref{fig:overview} shows an overview of our proposed framework for TPU-based acceleration of explainable machine learning. For a specific ML task, we apply traditional training scheme to construct a well-trained model and corresponding input-output dataset. Then we build a corresponding distilled model, which is able to provide reasonable explanation for target model's behavior. In this work, we consider three major tasks to achieve fast model distillation. First, we perform \textit{task transformation} to map the model distillation problem to Fourier transform computation by utilizing the inherent property of matrix convolution (Section~\ref{trans}). Next, we develop two synergistic activities to accelerate the computation procedure of Fourier transform using TPUs. The first activity performs \textit{data decomposition} (Section~\ref{decomp}), where the complete computing task is split into multiple sub-tasks, and each sub-task can be executed by a TPU core without requiring any data exchange between TPU cores (sub-tasks). The other one exploits the TPU's inherent ability in \textit{parallel computation} (Section~\ref{paral}) to process multiple input-output pairs concurrently. Simultaneous execution of these two activities can provide significant improvement in acceleration  efficiency, which is demonstrated in our experimental evaluation (Section~\ref{exp}).

\subsection{Task Transformation}
\label{trans}
As described in Section~\ref{relwork}, model distillation is applied in our work to provide explanation for a pre-trained ML model. In this section, we demonstrate how to convert model distillation task into a matrix computation. Specifically, model distillation aims at distilling features learned by a complex and cumbersome ML model, and use a lightweight ``shadow'' model to mimic its input-output mapping behavior, which is called as \textit{distilled model}. To make the distilled model useful, two crucial requirements need to be satisfied.
\begin{itemize}
    \item \textit{Simplicity:} The distilled model should be lightweight and simple. Otherwise it is difficult for users to understand the behaviors of distilled model.
    \item \textit{Compatibility:} The type of distilled model should be compatible with the original one. For instance, use of a fully-connected network to approximate a convolution neural network would lead to loss of accuracy.  
\end{itemize}

To satisfy these requirements, our proposed solution consists of the following three steps: model specification, model computation and outcome interpretation. 

\textit{\textbf{Model Specification:}} In this work, a regression model is applied in order to satisfy the two requirements outlined above. Formally, given input data $\textbf{X}$, output $\textbf{Y}$, we need to find a matrix $\textbf{K}$ using Equation~\ref{eqn:spec}, where ``*'' denotes the matrix convolution.
\begin{equation}
\textbf{X} * \textbf{K} = \textbf{Y}
\label{eqn:spec}
\end{equation}

Intuitively, we are using a one-layer convolution network to approximate the original ML model. 
First, convolution is a linear-shift-invariant operation, which guarantees the distilled model to be sufficiently lightweight and transparent. Second, a large portion of ML models encountered in reality are mainly composed of convolution layers. Moreover, convolution approximation is a natural fit and is more accurate than naive approaches such as polynomial approximation. Under this scenario, the model computation task boils down to solving for the parameters in matrix ${\bf K}$.

\textit{\textbf{Model Computation:}} To solve for ${\bf K}$, one key observation is that we can apply Fourier transformation on both sides of Equation~\ref{eqn:spec}, and by discrete convolution theorem, it gives
\begin{equation}
\begin{split}
\textbf{X} * \textbf{K} &= \textbf{Y}\\
\mathscr{F}(\textbf{X} * \textbf{K}) &= \mathscr{F}(\textbf{Y})\\
\mathscr{F}(\textbf{X}) \circ \mathscr{F}(\textbf{K}) &= \mathscr{F}(\textbf{Y})
\end{split}
\label{eqn:model}
\end{equation}
where $\circ$ is the Hadamard product. Therefore, the solution is given by this formula:
\vspace{-0.05in}
\begin{equation}
\textbf{K} = \mathscr{F}^{-1}(\mathscr{F}(\textbf{Y})/\mathscr{F}(\textbf{X}))
\label{eqn:hadamard}
\vspace{-0.05in}
\end{equation}

\textit{\textbf{Outcome Interpretation:}} The primary goal of explainable ML is to measure how each input feature contributes to the output value. Once $\textbf{K}$ is obtained, the contribution of each feature can be viewed in an indirect way -- consider a scenario where we remove this component from the original input, and let it pass through the distilled model again to produce a ``perturbed'' result. Then by calculating the difference between the original and newly generated outputs, the impact of the key feature on the output can be quantized. The intuition behind the assumption is that hiding important features are more likely to cause considerable changes to the model output. 
Formally, assume that the input is $\textbf{X} = [\textbf{x}_1, \textbf{x}_2, ...,\textbf{x}_{i-1}, \textbf{x}_{i}, \textbf{x}_{i+1}..., \textbf{x}_d]$. We define the contribution factor of $\textbf{x}_i$ as

\vspace{-0.25in}
\begin{equation}
\begin{split}
con(\textbf{x}_i) \triangleq \textbf{Y} - \textbf{X}^\prime * \textbf{K}
\end{split}
\label{eqn:contribution}
\vspace{-0.4in}
\end{equation}
where $\textbf{X}^\prime = [\textbf{x}_1, \textbf{x}_2, ...,\textbf{x}_{i-1}, \textbf{0}, \textbf{x}_{i+1}..., \textbf{x}_d]$, which is nothing but removing the target component from the original input.

As we can see, the original model distillation task has been fully converted into a matrix computation problem, which consists of matrix convolution, point-wise division and Fourier transform only. The first two types of operations are inherently accelerated by TPU's built-in structure~\cite{intro}. The next section describes the details for accelerating Fourier transform.

\subsection{Data Decomposition in Fourier Transform}
\label{decomp}
In this section, we demonstrate how to apply data decomposition to disentangle Fourier transform computation, and further utilize TPU's computation resource to significantly accelerate the computing process. The general form of a 2-D Discrete Fourier Transform (DFT) applied on an $M\times N$ signal is defined as:
\vspace{-0.05in}
\begin{equation} \label{eq:1}
X[k,l] = \frac{1}{\sqrt{MN}}\sum_{n=0}^{N-1} \left[ \sum_{m=0}^{M-1} x[m,n] e^{-j2\pi\frac{mk}{M}} \right] e^{-j2\pi\frac{nl}{N}}
\vspace{-0.05in}
\end{equation}
where $k = 0, ..., M-1$ , $l = 0, ..., N-1$. 

\noindent If we define intermediate signal $X'$ such that
\vspace{-0.05in}
\begin{equation} \label{eq:2}
X'[k,n] \triangleq \frac{1}{\sqrt{M}} \sum_{m=0}^{M-1} x[m,n] e^{-j2\pi\frac{mk}{M}} 
\vspace{-0.05in}
\end{equation}
and plug into Equation~\ref{eq:1}, we have
\vspace{-0.05in}
\begin{equation} \label{eq:3}
X[k,l] = \frac{1}{\sqrt{N}}\sum_{n=0}^{N-1}X'[k,n] e^{-j2\pi\frac{nl}{N}}
\vspace{-0.05in}
\end{equation}

Notice the similarity between Equation~\ref{eq:2} and the definition of 1-D Fourier transform applied on a $M$-length vector:
\vspace{-0.05in}
\begin{equation}
X[k] = \frac{1}{\sqrt{M}} \sum_{m=0}^{M-1} x[m] e^{-j2\pi\frac{mk}{M}} 
\vspace{-0.05in}
\end{equation}

If we treat $n$ as a fixed parameter, then application of Equation~\ref{eq:2} is equivalent to performing a 1-D Fourier transform on the $n$-th column of the original input $M \times N$ matrix. Note that for 1-D Fourier transform, it can always be written as a product of input vector and  Fourier transform matrix. Therefore, 
we can rewrite Equation~\ref{eq:2} as: 
\vspace{-0.05in}
\begin{equation}
X'[k,n]={\bf W}_M \cdot x[m,n]
\vspace{-0.05in}
\end{equation}
where ${\bf W}_M$ is the $M \times M$ Fourier transform matrix. By varying $n$ from 1 to $N-1$ and showing results side by side, we get:
\vspace{-0.05in}
\begin{equation}    
\label{eq:6}
X'=\left[X'[k,0],\cdots,X'[k,N-1]\right] = {\bf W}_M \cdot x
\vspace{-0.05in}
\end{equation}

If we treat $k$ as a parameter and view the definition of $X'[k,n]$ as the 1-D Fourier transform with respect to the $k$-th row of input $x$, a similar expression can be obtained using the above derivation steps as:
\vspace{-0.1in}
\begin{equation}
X = X' \cdot {\bf W}_N 
\vspace{-0.05in}
\end{equation} 
where ${\bf W}_N$ is the $N \times N$ Fourier transform matrix. Using Equation~\ref{eq:6}, the final expression of X can be written as:
\vspace{-0.05in}
\begin{equation}
X = ({\bf W}_M\cdot x) \cdot {\bf W}_N 
\vspace{-0.05in}
\end{equation}

This transformed expression indicates that a 2-D Fourier transform can be achieved in a two-stage manner. First, transform all the rows of $x$ to obtain intermediate result $X'$. Second, transform all the columns of the resulting matrix $X'$. An important observation is that the required computation for each row/column are completely \textbf{independent}. This implies that in TPU-based implementation, we can always split the computation process into sub-threads. Given $p$ individual TPU cores involved and a $M \times N$ matrix as input, every core is assigned at most $\frac{max\{M,N\}}{p}$ 1-D Fourier transforms workload and can execute in parallel.
Our analysis reveals that merging the results afterwards exactly matches the desired 2-D Fourier transform result. Algorithm~\ref{alg:alg1} outlines the major steps in data decomposition.

\begin{algorithm}[h]
\SetKwInOut{Input}{Input}
\SetKwInOut{Output}{Output}
\Input{$M \times N$ matrix $x$, number of TPU cores $p$}
\Output{2D Fourier Transform result $X$} 
 Initialize each TPU core $c_1, c_2,...c_p$\\
 $X = {\bf 0}$\\
 \For{each $i \in [0, ..., p-1]$}{
 Split $M/p$ rows $x_i$ from $x$\\
 $X'_i$= $Execute(c_i, x_i)$
 }
 Merge Results: $X' = [X'_1, X'_2, ..., X'_{p}]^T$\\
 \For{each $j \in [0, ..., p-1]$}{
 Split $N/p$ columns $x'_j$ from $X'$\\
 $X_j$= $Execute(c_j, x'_j)$
 }
 Merge Results: $X = [X_1, X_2, ..., X_{p}]$\\
 \textbf{return} $X$\\
 $   $\\
 \textbf{procedure} $Execute(c_i, x_i)$\\
   $res = {\bf 0}$ \\
   \For{each $r \in x_i$}{
   $r'= \mathscr{F}(r)$\\
   }
   $res = merge(res, r)$\\
   \textbf{return} $res$\\
 \textbf{endprocedure}\\

\caption{Acceleration of Fourier Transform}
\label{alg:alg1}
\end{algorithm}

\subsection{Parallel Computation of Multiple Inputs}
\label{paral}
In addition to exploiting TPU to accelerate Fourier transform, we make use of TPU's parallel computing ability to further improve the time efficiency. Notice in the training phase, multiple inputs will be fed into the model to generate corresponding outputs. The above data decomposition technique is applied on each individual input such that the computation cost is distributed among several TPU cores. Extending from single to multiple input is straightforward and only requires one-step further utilization of parallel computation.

\begin{figure}[htbp]
\centering
\vspace{-0.1in}
\includegraphics[scale=0.3]{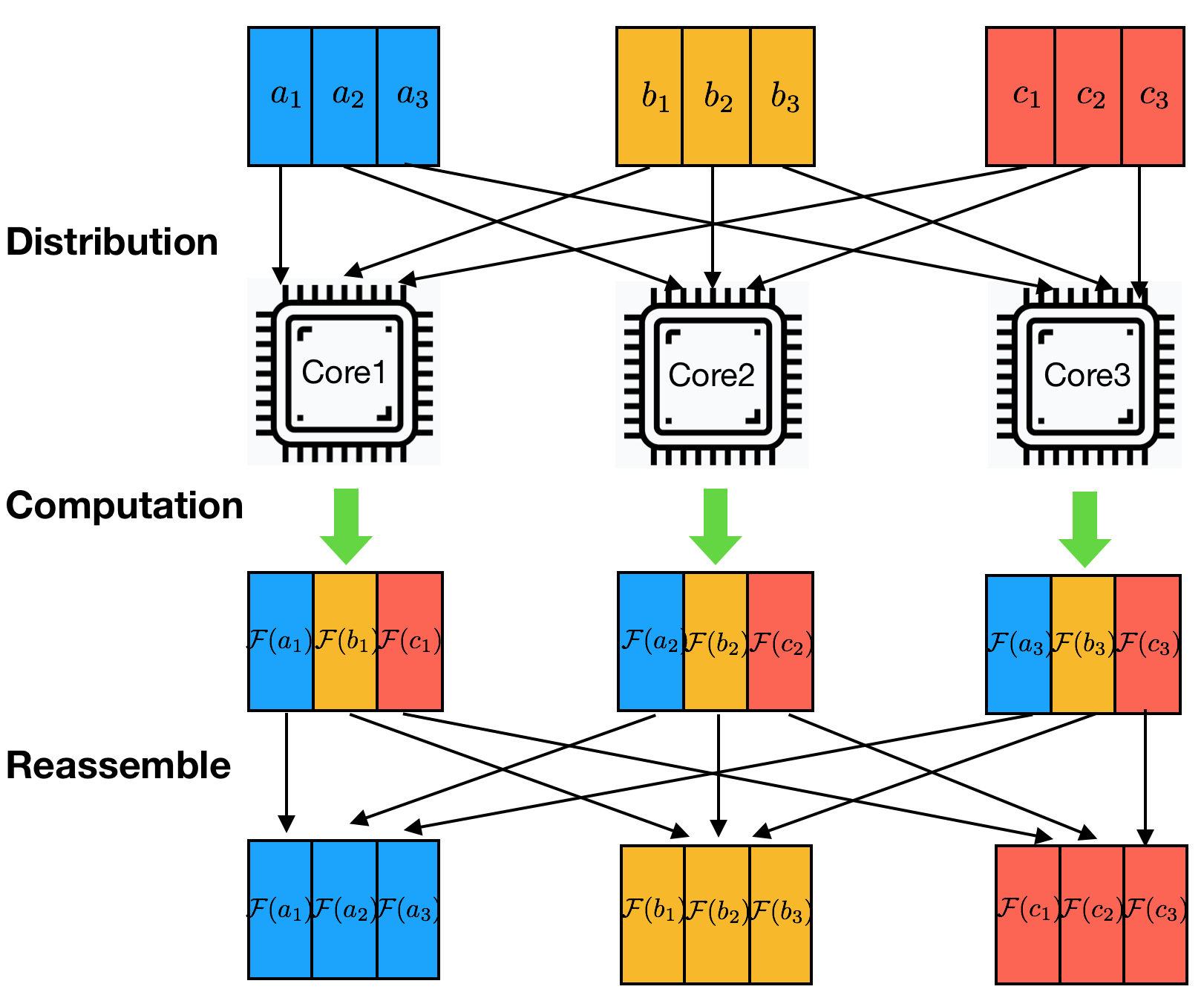}
\vspace{-0.05in}
\caption{An example of parallel computing in TPU. Each input is separated into pieces for  multiple cores to run in parallel. The outputs are reassembled to obtain the desired results.}
\label{fig:parallel}
\vspace{-0.2in}
\end{figure}

 An illustrative example is shown in Figure~\ref{fig:parallel} where the goal is to perform 1-D Fourier transform on each column of three input matrices. First, each input matrix is segmented into pieces and each core obtains a slice of them. Next, each piece is assigned to a TPU core to perform the Fourier transform. During computation, an internal  table is utilized to keep track of the distribution to guide the process of reassembling. 
 
 In terms of matrix multiplication, the framework is exactly the same except the fact that \textit{block matrix multiplication} is applied. Original matrices are partitioned into small blocks, then by performing multiplication between blocks and merging afterwards, we achieve same-level of parallel computing efficiency. Due to the data decomposition step  applied in Section~\ref{decomp}, the whole computing procedure contains Fourier transform, matrix multiplication and point-wise division only, which indicates parallelism is maintained across the entire computation process.
 
 In this work, the communication among TPU cores is implemented with $tf.cross\_replica\_sum$ and is required at every iteration of reassembly process to compute the summation of the partial matrices across the cores. The proposed data decomposition and the parallel implementation not only efficiently utilizes TPU’s strength in matrix multiplication but also requires minimal communication time, which leads to drastic improvement in acceleration performance.
\section{Experimental Evaluation}\label{exp}
We evaluated the effectiveness of our TPU-based framework in accelerating explainable machine learning models.

\subsection{Experimental Setup}

Experiments were conducted on a host machine with Intel i7 3.70GHz CPU, equipped with an external NVIDIA GeForce GTX 1080 GPU. We utilize Google's Colab platform to access Google Cloud TPU service. In our evaluation, we used TPUv2 with 64 GB High Bandwidth Memory (HBM), and 128 TPU cores. We developed code using Python~\cite{Python} for model training and PyTorch 1.6.0~\cite{Pytorch} as the machine learning library. Two benchmarks are considered in our experiment.

\begin{enumerate}
    \item The VGG19~\cite{VGG19} classifier for \textit{CIFAR-100}~\cite{CIFAR} image classification.
    \item The ResNet50~\cite{ResNet} network for \textit{MIRAI}~\cite{MalwareDataset} malware detection.
\end{enumerate}

We have used the following three hardware configurations to highlight the importance of our proposed TPU-based acceleration approach. Meanwhile, to address the compatibility of proposed optimization approach (Algorithm~\ref{alg:alg1}) and TPU, same optimization methods (data decomposition and parallel computing) are also deployed on CPU and GPU:
\begin{enumerate}
\item \textbf{CPU}: Ordinary execution with CPU, which is considered as baseline method.
\item \textbf{GPU}: Model training and outcome interpretation are deployed on the external NVIDIA GPU, which is considered as state-of-the-art ML acceleration technique.
\item \textbf{TPU}: Our proposed approach  with fast implementation of explainable machine learning procedure.
\end{enumerate}

The model training process consists of 500 epochs in total, with a mini-batch size of 128. 
As for result evaluation, we first evaluated classification performance by reporting ML models' classification accuracy and execution time. Next, we report the average time for completing outcome interpretation step for each configuration. Finally, we present the effectiveness of proposed method in interpreting classification results.

\subsection{Comparison of Accuracy and Classification Time}
Table~\ref{tb_1} compares the classification time and accuracy. Each row represents for a specific model structure trained with corresponding hardware configuration. For both training time and testing time, each entry represents time cost of 10 epochs on average. As we can see, with sufficient number of training epochs, all methods obtain reasonable classification accuracy. However, when it comes to time-efficiency, CPU-based approach lags far behind the other two, which achieved the slowest speed. GPU provides accelerated performance, but our TPU-based acceleration provides 65x and 25.7x speedup compared to CPU and GPU based methods, respectively, on VGG19. When it comes to ResNet50, high speedup (44.5x and 23.9x) are also obtained. Such a drastic improvement will also lead to significant energy savings by our proposed approach compared to CPU and GPU-based methods. 

\begin{table*}[h]
    \centering
    \caption{Comparison of accuracy and classification time for various benchmarks}
    \vspace{-0.1in}
    \label{tb_1}
    \begin{tabular}{|c|c|c|c|c|c|c|c|c|c|c|c|}
        \hline
        & \multicolumn{3}{c|}{CPU-based Acceleration} 
        & \multicolumn{3}{c|}{GPU-based Acceleration} 
        & \multicolumn{5}{c|}{TPU-based Acceleration (Proposed Approach)}\\
        \hline
        bench  & Accuracy& Training- & Testing-  & Accuracy & Training- & Testing- & Accuracy & Training- & Testing- & Speedup. & Speedup.\\
        &  (\%) &  time(s) & time(s) &  (\%) & time(s) & time(s) & (\%) & time(s) & time(s) & /CPU & /GPU\\
        \hline
        VGG19  & 94.06 & 24.2 & 10.9  & 92.08 & 8.1 & 5.8  & 96.37 & 0.4 & 0.14 & 65x & 25.7x\\
        \hline
        ResNet50  & 78.99 & 176.2 & 129.8 &  86.87 & 109.7 & 55.0  & 87.47 & 4.3 & 2.60 & 44.5x & 23.9x\\
        \hline
        {\bf Average} &  {\bf 86.52} & {\bf 100.2} & {\bf 70.35} & {\bf 89.47} & {\bf 58.9} & {\bf 30.4} &  {\bf 91.92} & {\bf 2.35} & {\bf 1.37} & {\bf 54.7x} & {\bf 24.8x}\\
        \hline
    \end{tabular}
\end{table*}



\subsection{Scalability of Outcome Interpretation}
In this section, we demonstrate the efficiency of our proposed method on explaining ML models.
The average time for performing outcome interpretation for every 10 input-output pairs is presented in Table~\ref{tb_2}. The VGG19 result demonstrates that the proposed method is 36.2x and 11x faster than CPU and GPU based approaches, respectively. As expected, the improvements are even higher in case of ResNet50, where 39.5x and 13.6x speedup obtained over CPU and GPU based approaches, respectively.

\begin{table}[ht]
    \centering
    \vspace{-0.05in}
    \caption{Average time (seconds) for outcome interpretation} 
    \label{tab:time}
    \begin{tabular}{|c|c|c|c|c|c|}
        \hline
        Model & CPU & GPU  & TPU & Impro./CPU & Impro./GPU \\
        \hline
        VGG19 & 550.7 & 168  & 15.2s & 36.2x & 11x\\
        \hline
        ResNet50 & 1456.1 & 502  & 36.8s & 39.5x & 13.6x\\
        \hline
        {\bf Average} & {\bf 1003.4} &{\bf 335}  & {\bf 26.0s} & {\bf 38.6x} & {\bf 12.8x} \\
        \hline
    \end{tabular}
    \label{tb_2}
\end{table}

\begin{figure}[htbp]
\vspace{-0.2in}
\centering
\includegraphics[scale=0.15]{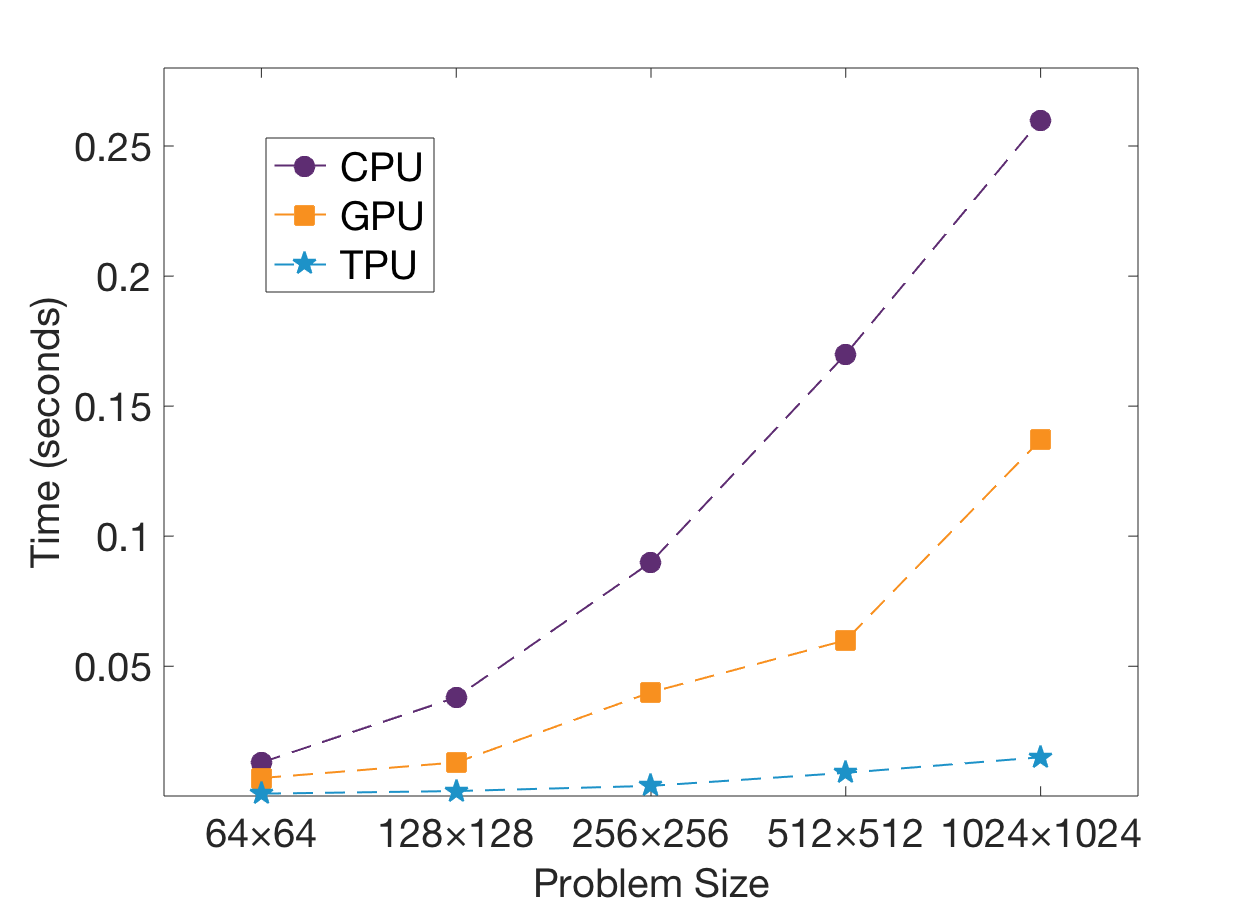}
\caption{Scalability of three methods}
\label{fig:sz}
\vspace{-0.2in}
\end{figure}

Aside from the overall efficiency analysis presented above, we also randomly select several matrices with varying sizes and compare time efficiency in Figure~\ref{fig:sz}. It is expected that the time will increase with the increase in the size of the matrices. Figure~\ref{fig:sz} demonstrates that our proposed approach is scalable compared to the other two methods. There are two reasons for the scalability of our approach. (1) Our approach utilized data decomposition technique to break larger matrices into smaller sub-matrices. (2) Another dimension of improvement comes from the fact that these smaller sub-matrices are distributed across multiple MXU inside each TPU core. This drastically reduces the bandwidth requirement during the computation and leads to a significant improvement in computation speed. For matrices in the size of 1024 × 1024, proposed method is more than 30x faster than the baseline method. This indicates that for training and outcome interpretation on large-scale neural networks (tens of thousands of matrix operations involved), our proposed method can save hours of computation time, which will lead to significant energy benefit.

\subsection{Outcome Interpretation of Classification Results}
Our method not only achieved faster classification, but also provides effective explanation of classification results. We have evaluated outcome interpretation in a wide variety of scenarios. In this section, we provide two examples from two different domains (image classification and malware detection).

\begin{figure}[h]
\vspace{-0.15in}
\centering
\includegraphics[scale =0.27]{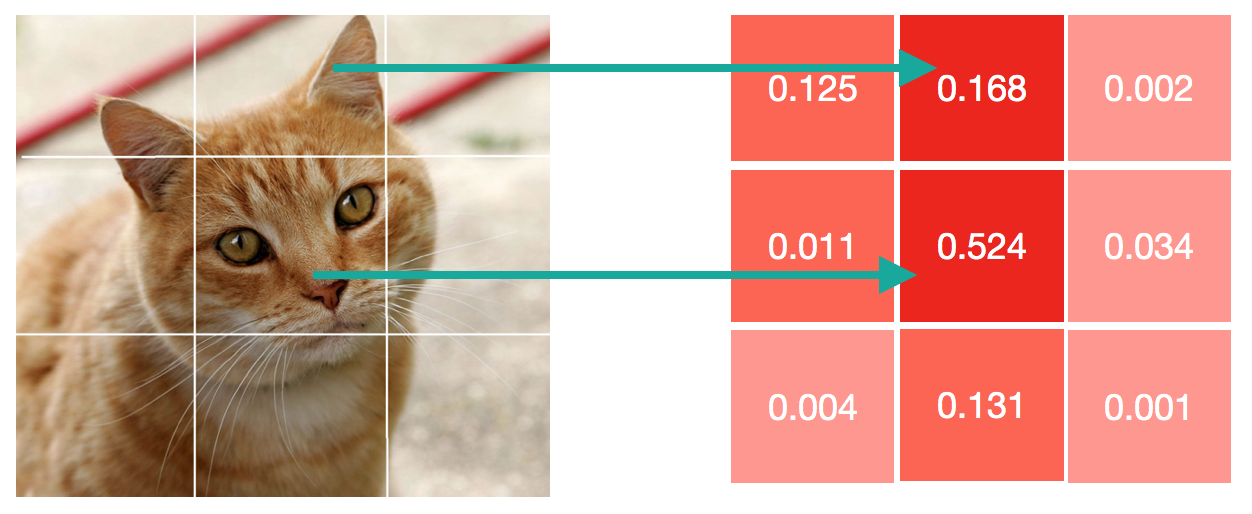}
\vspace{-0.1in}
\caption{Interpretation of CIFAR image's classification}
\vspace{-0.1in}
\label{fig:interp2} 
\end{figure}

Figure~\ref{fig:interp2} shows an example of interpreting the classification results for a picture from CIFAR-100 dataset. We segmented the given image into square sub-blocks, and the explainable ML framework is applied to compute contribution factor of each individual block towards the classifier's output, so that it can illustrate what part is crucial for the classifier to distinguish it from other categories. In the given picture, the cat's face (central block) and ear (mid-up block) are the keys to be recognized as `cat'.

\begin{figure}[h]
\vspace{-0.15in}
\centering
\includegraphics[scale =0.31]{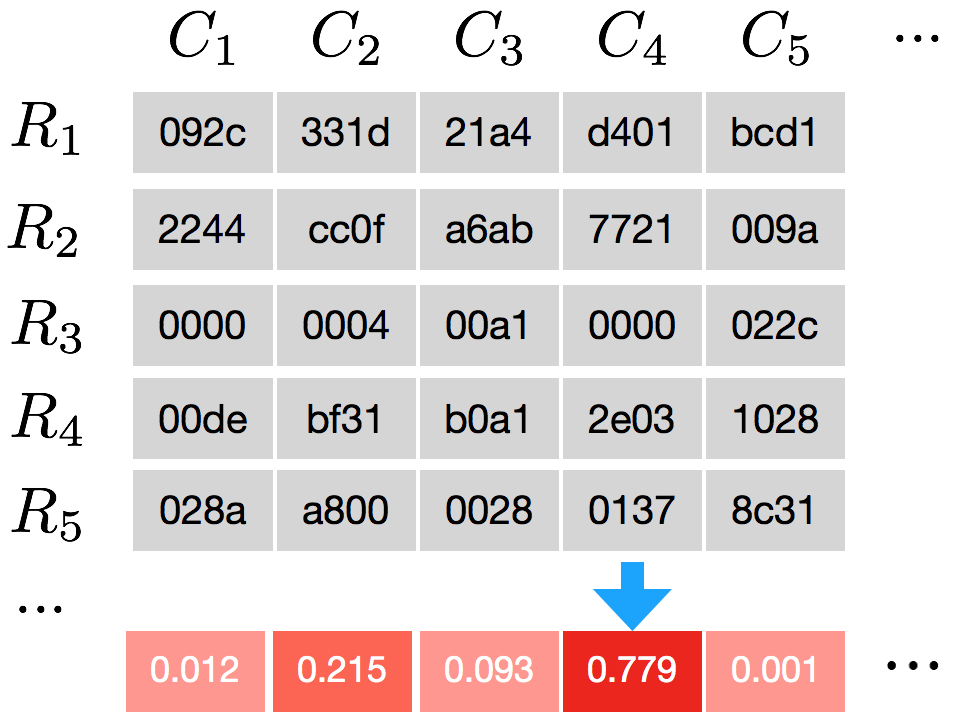}
\vspace{-0.1in}
\caption{Interpretation of MIRAI malware traced signals}
\label{fig:interp1} 
\vspace{-0.1in}
\end{figure}

Let us consider an example on malware detection from ResNet50. The ML-based detector receives running data of MIRAI malware \cite{MalwareDataset} as input in the format of a trace table, where each row represents the hex values in a register in specific clock cycles (each column represents a specific clock cycle). Figure~\ref{fig:interp1} shows a snapshot of the trace table.

 Our proposed method computed the corresponding contribution factor of each clock cycle towards the output using model distillation. Contribution factors are shown as weights in the last (colored) row. Clearly we can see that the weight of $C_2$ is significantly larger than the others. By tracing the execution, it has been shown that  $C_2$ corresponds to the timestamp of assigning value to the variable ``\textit{ATTACK\_VECTOR}" in Mirai. This variable records the identity of attack modes, based on which the bot takes relative actions to perform either a UDP attack or DNS attack. This attack-mode flag is the most important feature of a majority of malware bot programs, and our proposed method successfully extracted it from the traces to illustrate the reason for classifying it as a malware. This interpretation will not only provide confidence in malware detection but also helps in malware localization.

\section{Conclusion}\label{conclude}
While machine learning techniques are popular in many domains, they have two major limitations in real-world applications: long running time and lack of transparency. In this paper, we address these fundamental bottlenecks using TPU-based hardware acceleration of explainable machine learning. While acceleration improves the classification time, model distillation provides the explainability (transparency) of the classification results. Our proposed method made two important contributions. First, it effectively converts the model distillation problem to linear algebra computation. As a result, it is able to fully exploit TPU’s inherent ability in computing ultra-fast matrix operations. Next, it enables parallel computing by performing data decomposition to break a large matrix into multiple small matrices. Experimental results on a diverse set of benchmarks demonstrated that our approach is scalable and outperforms state-of-the-art acceleration techniques for explainable machine learning. Specifically, our TPU-based acceleration provides drastic improvement in classification time (25x on average) as well as interpretation time (13x on average) for both image classification and malware detection benchmarks over GPU-based acceleration. 


\bibliographystyle{IEEEtran}
\bibliography{zhixin.bib}

\end{document}